\documentclass{article}

     \usepackage[final,nonatbib]{tackling_climate_workshop_style}

\usepackage[utf8]{inputenc} 
\usepackage[T1]{fontenc}    
\usepackage{hyperref}       
\usepackage{url}            
\usepackage{booktabs}       
\usepackage{amsfonts}       
\usepackage{nicefrac}       
\usepackage{microtype}      
\usepackage{wrapfig}
\usepackage{graphicx}
\usepackage{xcolor}

\title{Can We Reliably Improve the Robustness to Image Acquisition of Remote Sensing of PV Systems?}

\author{%
Gabriel Kasmi$^{1,2}$ \quad Laurent Dubus$^{2,3}$ \quad Yves-Marie Saint Drenan$^1$\quad
Philippe Blanc$^{1}$\\
$^1$MINES Paris, Université PSL Centre Observation Impacts Energie (O.I.E.)\\
$^2$RTE France \\
$^3$WEMC (World Energy \& Meteorology Council, UK) \\
$^1$\texttt{\{firstname.lastname\}@minesparis.psl.eu}\\
$^2$\texttt{\{firstname.lastname\}@rte-france.com}\\
}

\begin{document}

\maketitle

\begin{abstract}
Photovoltaic (PV) energy is crucial for the decarbonization of energy systems. Due to the lack of centralized data, remote sensing of rooftop PV installations is the best option to monitor the evolution of the rooftop PV installed fleet at a regional scale. However, current techniques lack reliability and are notably sensitive to shifts in the acquisition conditions. To overcome this, we leverage the wavelet scale attribution method (WCAM) \cite{kasmi_assessment_2023}, which decomposes a model's prediction in the space-scale domain. The WCAM enables us to assess on which scales the representation of a PV model rests and provides insights to derive methods that improve the robustness to acquisition conditions, thus increasing trust in deep learning systems to encourage their use for the safe integration of clean energy in electric systems.
\end{abstract}

\section{Introduction}

Photovoltaic (PV) energy grows rapidly and is crucial for the decarbonization of electric systems \cite{haegel_terawatt-scale_2017}. The rapid growth of rooftop PV makes the estimation of the global PV installed capacity challenging as centralized data is often lacking \cite{kasmi_towards_2022}. Remote sensing rooftop PV on orthoimagery with computer vision models is a promising solution for mapping rooftop PV installations. However, current approaches lack reliability and generalize poorly from one image provider to the other \cite{wang_poor_2017,hu_what_2022}. Improving generalizability and reliability requires improving the robustness to acquisition conditions and proposing methods to assess the reliability of the decision process of the PV classifiers \cite{de_jong_monitoring_2020}.

Deep learning-based pipelines became the standard method for remote sensing PV systems. DeepSolar \cite{yu_deepsolar_2018} paved the way for country-wide mapping of PV systems using deep learning and overhead imagery. While several works discuss the poor generalizability of current methods \cite{wang_poor_2017,frimane_identifying_2023, hu_what_2022}, these works do not tackle the robustness to varying acquisition conditions, which can be assimilated to image corruptions \cite{hendrycks_benchmarking_2019}. 



In this work, we analyze the robustness to heterogeneous acquisition conditions of models for remote sensing of PV installations using the wavelet scale attribution method (WCAM, \cite{kasmi_assessment_2023}). The WCAM assesses the reliability of a model's decision by decomposing it into the scale-space domain. We analyze the model's sensitivity to acquisition conditions using the WCAM and derive a principled method for improving the robustness to these acquisition conditions. Our work shows that the WCAM provides a finer understanding of what the model sees as a PV panel and guides us to improve the robustness to acquisition conditions. By improving the reliability and robustness of deep learning models for rooftop PV mapping, we aim to facilitate the mapping and, thus, the integration into the electric grid of rooftop PV.

\section{Related works}

\paragraph{Remote sensing of PV installations} Many works leveraged overhead imagery and deep learning methods to map PV installations \cite{malof_automatic_2015,malof_automatic_2016,golovko_development_2018,yuan_large-scale_2016}. The DeepSolar \cite{yu_deepsolar_2018} method marked a significant milestone with mapping distributed and utility-scale installations over the continental United States using state-of-the-art deep learning models. Many works built on DeepSolar to map regions or countries, especially in Europe \cite{kausika_geoai_2021,arnaudo_comparative_2023,kasmi_towards_2022,frimane_identifying_2023,mayer_3d-pv-locator_2022,mayer_deepsolar_2020}. 
However, current methods cannot be transposed from one region to another without incurring accuracy drops, thus limiting their practical usability \cite{hu_what_2022} due to a lack of reliability of the generated data \cite{de_jong_monitoring_2020}. To address this gap, we propose to study and mitigate the impact of acquisition conditions, ubiquitous with overhead imagery, which prevents reusing trained models for registry updates. 

\paragraph{Sensitivity to distribution shifts} The sensitivity to distribution shifts \cite{koh_understanding_2020} prevents from using pre-trained models without further training, whether temporally or spatially, i.e., it limits their ability to generalize \cite{de_jong_monitoring_2020,malof_mapping_2019}. Some works empirically discussed this issue \cite{wang_poor_2017} and argued that the generalization ability depended on how hard to recognize the PV panels are. However, no work properly disentangled the effect of each source of variability identified by \cite{tuia_domain_2016}: geographical conditions, varying acquisition conditions, and the ground sampling distance (GSD). 

\paragraph{Frequency-centric explanations} A line of works aimed at explaining the behavior of neural networks through the lenses of frequency analysis. Several works showed that convolutional neural networks (CNNs) are biased towards high frequencies \cite{wang_high-frequency_2020,yin_fourier_2020} and that robust methods tend to limit this bias \cite{zhang_range_2022,chen_rethinking_2022}. Other works highlighted a so-called spectral bias \cite{rahaman_spectral_2019,xu_frequency_2020,jo_measuring_2017}, showing that CNNs learn the input frequencies from the lowest to the highest. More recently, using wavelet transforms, \cite{kasmi_assessment_2023} expanded attribution methods from the pixel to the space-scale (wavelet) domain. This work connects the fields of interpretability and robustness and enables understanding {\it what} models see on images. It has not yet been applied to orthoimagery, where scales are explicitly indexed.

\section{Data and methods}

\subsection{Data}

We consider the crowdsourced training dataset {\it Base de données d'apprentissage profond photovoltaique} (BDAPPV, \cite{kasmi_crowdsourced_2023}). This dataset contains annotated images of 28,000 PV panels in France and neighboring countries. This dataset also proposes annotations of images that depict the same PV panel but from two different image providers: images coming from the Google Earth Engine (hereafter referred to as "Google") \cite{gorelick_google_2017} and from the IGN, the French public operator for geographic information. We have double annotations for more than 8,000 PV systems. It allows us to assess the impact of the acquisition conditions as the only change factor between two images is the varying acquisition condition: the semantic content (the PV panel and its surroundings) remains unchanged. The native ground sampling distance (GSD) of Google images is 10 cm/pixel and 20 cm/pixel for IGN images. We define the acquisition conditions as the properties of the technical infrastructure (airborne or spaceborne, camera type, image quantization, and postprocessing) and the atmospheric and meteorological conditions the day the image was taken. \autoref{fig:examples} in the appendix \ref{sec:figures} presents images samples of the BDAPPV dataset. 

\subsection{Methods}

\subsubsection{Identifying where the sensitivity to distribution shifts comes from}

\paragraph{Empirical framework} BDAPPV features images of the same installations from two providers and records the crude location of the PV installations. Using this information, we can define three test cases to disentangle the distribution shifts that occur with remote sensing data: the resolution, the acquisition conditions, and the geographical variability. We train a ResNet-50 model \cite{he_deep_2016} on Google images downsampled at 20cm/pixel of resolution and evaluate it on three datasets: a dataset with Google images at their native 10cm/pixel resolution ("Google 10 cm/pixel"), the IGN images with a native 20cm/pixel resolution ("IGN") and Google images downsampled at 20 cm/pixel located outside of France ("Google OOD\footnote{OOD: out-of-distribution.}"). We add the test set to record the test accuracy without distribution shift ("Google baseline"). We only do random crops, rotations, and ImageNet normalizations during training. \autoref{fig:test-images} in appendix \ref{sec:figures} presents examples of images seen during training and test. 


\subsubsection{Data augmentations for improving the robustness to acquisition conditions}

\paragraph{Benchmarking current approaches} The literature on robustness to image corruptions \cite{hendrycks_benchmarking_2019} proposed numerous data augmentation methods to improve the robustness of classification models to image corruptions\cite{hendrycks_augmix_2020,hendrycks_pixmix_2022,cubuk_randaugment_2019,cubuk_autoaugment_2019,geirhos_imagenet-trained_2023}. We consider the well-established AugMix method \cite{hendrycks_augmix_2020} and the recently-proposed RandAugment \cite{cubuk_randaugment_2019} and AutoAugment \cite{cubuk_autoaugment_2019} methods. These methods apply a random composition of perturbations to images during training to learn an invariance against these perturbations. We do not consider the case of training from multiple sources as our setting is that we wish to generalize to unseen images (either temporally or spatially, so we cannot incorporate knowledge about these images).

\paragraph{Lowering the reliance on high-frequency components} Since we know that varying acquisition conditions mainly alter high-frequency components, we introduce two data augmentation techniques that aim at reducing the reliance on high-frequency components: Gaussian blurring ("Blurring") and Blurring + wavelet perturbation (WP). Blurring consists of a fixed image blur, while the Blurring + WP also perturbs the wavelet coefficients of the image to force the model to rely on several rather than one scale for prediction. We refer the reader to the appendix \ref{sec:augmentation-plot} for more details on the data augmentation strategies and a review of the hyperparameters. \autoref{fig:augmentations} in appendix \ref{sec:plot-augmentation} illustrates the effect of the different data augmentation techniques. We compare this approach with a baseline without augmentations ("ERM") and existing data augmentation techniques.

\subsubsection{Understanding the sensitivity to acquisition conditions with the Wavelet sCale Attribution Method (WCAM)}

Attribution \cite{simonyan_very_2015,selvaraju_grad-cam_2020,petsiuk_rise_2018} indicates the important regions for prediction, i.e., decomposes the prediction in the pixel (spatial) domain. The WCAM \cite{kasmi_assessment_2023} generalizes attribution to the wavelet (space-scale domain). The WCAM provides us with two pieces of information: where the model sees and what scale (i.e., frequency) it sees at this location. Therefore, we can see if a prediction relies on robust or fragile frequencies. Additionally, the decomposition of the prediction in terms of scales is interpretable, particularly in the case of orthoimagery. For example, on Google images, details at the 1-2 pixel scale correspond to physical objects with a size between 0.1 and 0.2 m on the ground. Thus, we know what the model sees as a panel; we can interpret it and assess whether it is sensitive to varying acquisition conditions. The decomposition brought by the WCAM enables the interpretation of the model's decision process. Appendix \ref{sec:read-wcam} provides additional background for reading WCAMs.

\section{Results}

\subsection{Acquisition conditions mainly explain the poor generalization of PV mapping algorithms}

\begin{table}[h]
\small
  \centering
  \vspace{0mm}\caption{\textbf{F1 Score} and decomposition in true positives, true negatives, false positives, and false negatives rates of the disentanglement of the distribution shift between the GSD (Google 10 cm/px), the geographical variability (Google OOD) and the acquisition conditions (IGN).}\label{tab:baseline-results} 
  \resizebox{\textwidth}{!}{\begin{tabular}{r c c c c c}
  \toprule
   & F1 Score ($\uparrow$) & True positives rate & True negatives rate & False positives rate & False negatives rate \\
   \midrule
    Google baseline &  0.98  & 0.99 & 0.98 & 0.02& 0.01 \\ 
    Google 10cm/px & {\color{orange}0.89} &  0.81 & 1.00 & 0.00& {\color{orange}0.19}\\
    Google OOD &0.98 &  0.99 & 0.98 & 0.02& 0.01\\
    IGN & {\color{red}0.46} & 0.32 & 0.95 & 0.03& {\color{red}0.68}\\

  \bottomrule
  \end{tabular}}
  \end{table}

\paragraph{Results} \autoref{tab:baseline-results} shows the results of the decomposition of the effect of distribution shifts into three components: resolution, acquisition conditions, and geographical shift. We can see that the F1 score drops the most when the model faces new acquisition conditions. The second most significant impact comes from the change in the ground sampling distance, but the performance drop remains relatively small compared to the effect of the acquisition conditions. In our framework, there is no evidence of an effect of the geographical variability once we isolate the effects of the acquisition conditions and ground sampling distance. This effect is probably underestimated, as images of our dataset that are not in France are near France. However, the effect of the acquisition conditions is sizeable enough to seek methods for addressing it.

 \begin{wrapfigure}{r}{0.5\textwidth}
\centering
\includegraphics[width=.5\textwidth]{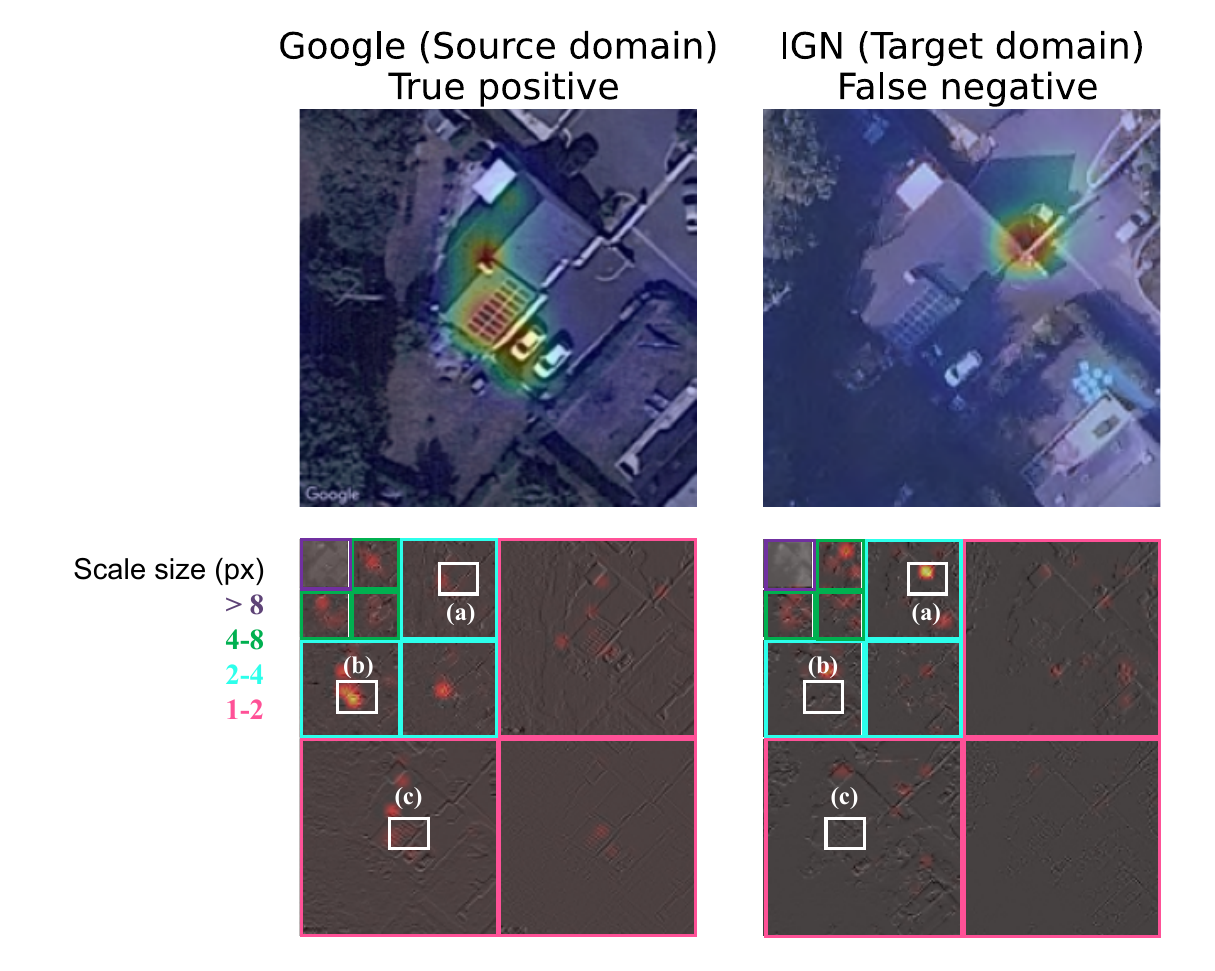}
\caption{Predictions on Google image (left, upper row) and IGN image (right, upper row) and associated WCAMs (bottom row, displayed with the same color scale). The brighter, the more important the highlighted region for the prediction}\label{fig:wcam-example}
\vspace{-6mm}
\end{wrapfigure}
 
 \paragraph{Mechanisms: when important factors disappear} Changing the provider (i.e., altering the acquisition conditions) alters the scales describing the image of PV panels. If the model relied on a scale no longer on the image, it could no longer recognize the PV panel. On \autoref{fig:wcam-example}, we can see that on Google, the important factor was the factor {\bf (b)} (on the leftmost image), which is no longer important on the IGN image (on the right). On the IGN image, the model instead relied on the factor {\bf (a)} as the factor {\bf (b)} is no longer visible. This change in the important factor (at the same scale in this example) seems to have driven the shift from predicting to not predicting the PV panel. Interestingly, we can see that the scales highlighted in {\bf (c)} are visible on both images but not important for the prediction in the IGN image: the model no longer "sees" these details. We refer the reader to the appendix \ref{sec:decomposition} for further guidance on interpreting a WCAM.


\subsection{Lowering the reliance on high frequencies improves generalization}

\paragraph{Blurring and wavelet perturbation improve accuracy} \autoref{tab:mitigation} reports the results of the evaluated data augmentation techniques to mitigate the effect of acquisition conditions. Augmentations that explicitly discard small scales (high frequencies) information perform the best. However, the blurring method sacrifices the recall (which drops to 0.6) to improve the F1 score. On \autoref{tab:mitigation}, this can be seen by the increase in false positives. Therefore, this method is unreliable for improving the robustness to acquisition conditions. On the other hand, adding wavelet perturbation yields improvements and outperforms existing approaches without sacrificing precision or recall. 

  \begin{table}[h]
\small
  \centering
  \vspace{0mm}\caption{\textbf{F1 Score} and decomposition in true positives, true negatives, false positives, and false negatives for models trained on Google with different mitigation strategies. All models are evaluated on the same test set, so we report the raw values rather than the rates. Evaluation on IGN images. The oracle corresponds to a model trained on IGN images with standard augmentations. Best results are {\bf bolded}.}\label{tab:mitigation}
  \resizebox{\textwidth}{!}{\begin{tabular}{r c c c c c}
  \toprule
   & F1 Score ($\uparrow$) & True positives & True negatives & False positives & False negatives \\
   \midrule
   Oracle & 0.88 & 1818 & 1992 & 428 & 83 \\
   \midrule
    ERM \cite{vapnik_nature_1999} & 0.44 & 566 & 2321 & 99 & 1335 \\
    AutoAugment \cite{cubuk_autoaugment_2019} & 0.46 & 598 & 2318 & 102 & 1303\\
    AugMix \cite{hendrycks_augmix_2020} & 0.48 & 624 & 2318 & 102 & 1277 \\
    RandAugment \cite{cubuk_randaugment_2019} & 0.51 & 707 & 2280 & 140 & 1194 \\
    Blurring & {\bf 0.74} & 1855 & 1196 & 1224 & 46\\
    Blurring + WP &  \underline{0.58} & 896 & 2114 & 306 & 1005\\
  \bottomrule
  \end{tabular}}
  \end{table}

\begin{figure}[h]
    \centering
    \includegraphics[width = .7\textwidth]{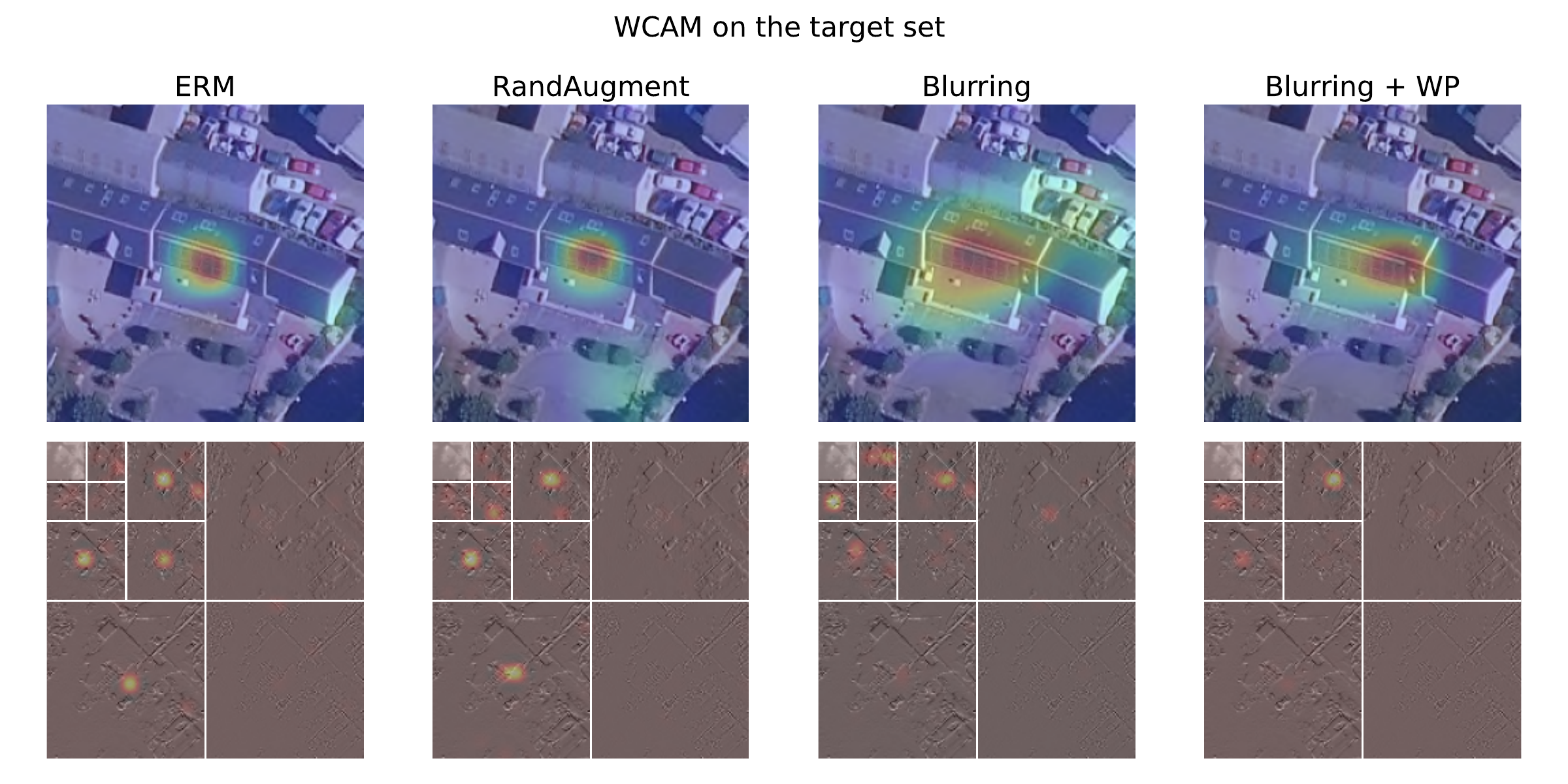}
    \caption{WCAMs on IGN of models trained on Google with different augmentation techniques.}
    \label{fig:wcam-strategies}
\end{figure}

\paragraph{Relying on consistent scales} \autoref{fig:wcam-strategies} compares the scales on which the best-performing methods rely. In our case, we want our models to rely on the largest scales (i.e., lowest frequencies) to entail robustness \cite{zhang_when_2022} against varying acquisition conditions. We can see that the blurring and wavelet perturbation enforces this property better than other data augmentation techniques. Indeed, the model relies on coarser scales (which are more robust) and on scales on which the ERM also relies. More generally, the WCAM lets us compare methods that perform quantitatively similarly.

\paragraph{On the choice of the training images} Our results show that lowering the reliance on high-frequency content in the image improves generalization. This content is located on the 10-20cm scale and only appears on Google images. In \autoref{tab:flipped-experiments}, we flip our experiment to study how a model trained on IGN images generalizes to Google images. Results show that the model trained on IGN generalizes better to the downscaled Google images than the opposite. This result further supports the idea that higher GSD is not necessarily better for good robustness to acquisition conditions.

  \begin{table}[h]
\small
  \centering
  \vspace{0mm}\caption{\textbf{F1 Score} and true positives, true negatives, false positives, and false negatives. Evaluation computed on the Google dataset. ERM was trained on Google and Oracle on IGN images.}\label{tab:flipped-experiments}
  \resizebox{\textwidth}{!}{\begin{tabular}{r c c c c c}
  \toprule
   
   & F1 Score ($\uparrow$) & True positives & True negatives & False positives & False negatives \\
   \midrule
    ERM \cite{vapnik_nature_1999} & 0.98 & 1891 & 2355 & 36 & 39 \\
    Oracle (ERM trained on IGN) &  0.91 & 1815 & 2127 & 264 & 115\\
  \bottomrule
  \end{tabular}}
  \end{table}

\section{Conclusions and future work}

We set up an experiment to disentangle the effects of heterogeneous acquisition conditions, geographical variability, and ground sampling distance on the generalization of deep neural networks to unseen data. Our results show that the sensitivity to acquisition conditions is the leading cause of poor generalization. To explain why models are sensitive to acquisition conditions, we leverage the wavelet scale attribution method (WCAM, \cite{kasmi_assessment_2023}). Acquisition conditions perturb the scales the model relied on to make a prediction. If these scales correspond to high frequencies, they are likely to be disrupted by the acquisition conditions. We show that models biased towards low frequencies are more robust to acquisition conditions. We design a data augmentation method that outperforms other methods to improve the robustness to varying acquisition conditions. More generally, models trained on images with a lower GSD generalize better.

\paragraph{Broader impact} Currently, transmission system operators (TSOs) lack quality data regarding rooftop PV installations \cite{kasmi_towards_2022}. The lack of information leads to imprecise estimations and forecasts of the overall PV power generation, which in a context of sustained growth of the PV installed capacity could increase the uncertainty and threaten the grid's stability \cite{pierro_impact_2022}. On the other hand, current methods for mapping rooftop PV installations lack reliability, owing to their poor generalization abilities beyond their training dataset \cite{de_jong_monitoring_2020}. This work addresses this gap and thus demonstrates that remote sensing of PV installations is a reliable way for TSOs to improve their knowledge regarding small-scale PV installations.

\paragraph{Future works} We wish to discuss further the conditions on the training images for good robustness to acquisition conditions. In particular, we plan to discuss the trade-off between the minimal GSD to {\it reliably} see PV panels \cite{li_understanding_2021} and a notion of image quality for the training data. 

\section{Acknowledgements}

This work is funded by RTE France, the French transmission system operator, and benefited from CIFRE funding from the ANRT. The authors gratefully acknowledge the support of this project. 

{\small
\bibliographystyle{plain}
\bibliography{references}
}

\newpage
\appendix

\section{Additional figures}\label{sec:figures}

\begin{figure}[h]
\centering
\includegraphics[width=0.7\textwidth]{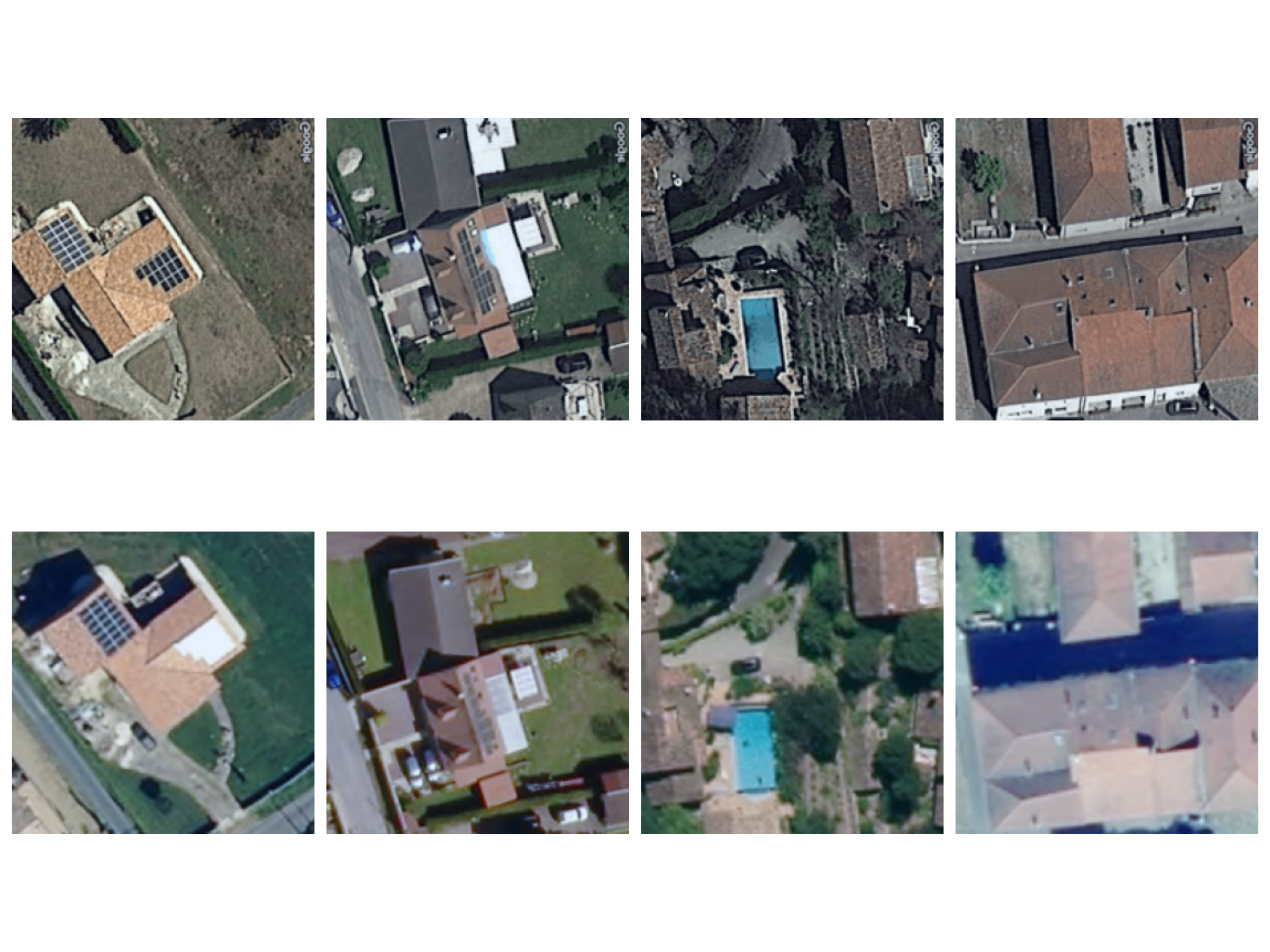}
\caption{Examples of test of the same PV panels but with different providers (Up Google, down: IGN).}\label{fig:examples}
\end{figure}

\paragraph{Examples of images from BDAPPV} \autoref{fig:examples} plots examples of images coming from BDAPPV. These images depict the same scene for two different providers.

\paragraph{Test sets}

\begin{figure}[h]
\centering
\includegraphics[width=0.7\textwidth]{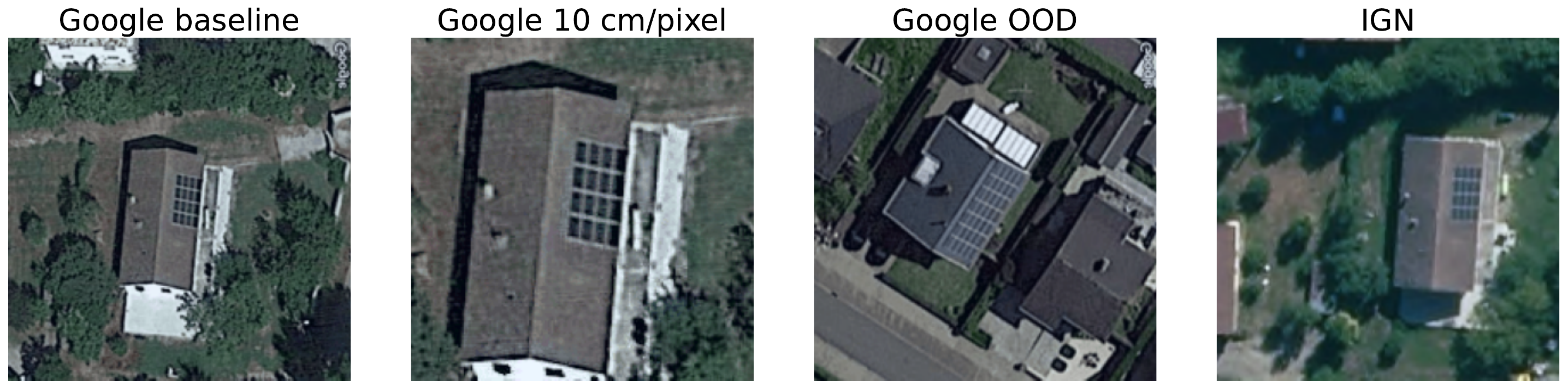}
\caption{Test images on which a model trained on Google images (downsampled to 20 cm/px of GSD) is evaluated. }\label{fig:test-images}
\end{figure}

\autoref{fig:test-images} plots examples of the different test images to disentangle the effects of distribution shifts. The baseline and IGN images represent the same panel at the same spatial resolution. The Google 10 cm/pixel depicts the same scene but with the native resolution of Google images. Finally, the OOD test set contains images located outside of France.

\section{Reading WCAMs}\label{sec:read-wcam}

\subsection{Identification of the important scales in the input image}

\begin{figure}[h]
    \centering
    \includegraphics[width = \textwidth]{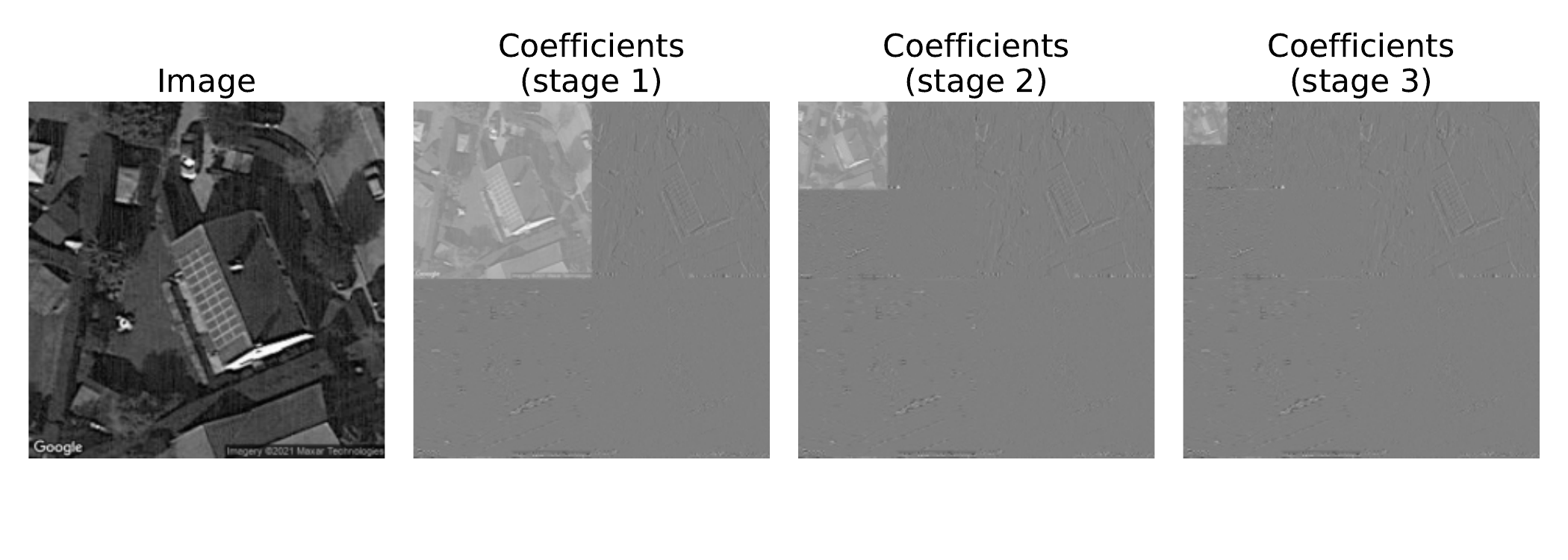}
    \caption{Example of a multilevel dyadic wavelet transform}
    \label{fig:wavelet-transforms}
\end{figure}

The content of this section comes from section 2.2. of \cite{kasmi_assessment_2023}. Our only addition is \autoref{fig:wavelet-transforms} and its comment (between "{\it \autoref{fig:wavelet-transforms} presents}" and "{\it larger than 8 pixels on the image.}"

\paragraph{Dyadic wavelet transform} A wavelet is an integrable function $\psi\in L^2 (\mathbb{R})$ with zero average, normalized and centered around 0. Unlike a sinewave, a wavelet is localized in space and in the Fourier domain. It implies that dilatations of this wavelet enable to scrutinize different frequencies (scales) while translations enable to scrutinize spatial location. To compute an image's (continuous) wavelet transform (CWT), one first defines a filter bank $ \mathcal{D}$ from the original wavelet $\psi$ with the scale factor $s$ and the 2D translation in space $u$. We have
\begin{equation}
        \mathcal{D} = \left\{
        \psi_{s,u}(x) = \frac{1}{\sqrt{s}}\psi\left(\frac{x-u}{s}\right)
    \right\}_{u\in\mathbb{R}^2,\;s\ge 0},
\end{equation}
where $\vert \mathcal{D}\vert =J$, and $J$ denotes the number of levels. The computation of the wavelet transform of a function $f\in L^2(\mathbb{R})$ at location $x$ and scale $s$ is given by
\begin{equation}\label{eq:wt}
    \mathcal{W}(f)(x,s) = \int_{-\infty}^{+\infty}
    f(u) \frac{1}{\sqrt{s}} \psi^*\left(\frac{x-u}{s}\right)
    \mathrm{d}u,
\end{equation}
which can be rewritten as a convolution \cite{mallat_wavelet_1999}. Computing the multilevel decomposition of $f$ requires applying \autoref{eq:wt} $J$ times with all dilated and translated wavelets of $\mathcal{D}$. \cite{mallat_theory_1989} showed that one could implement the multilevel dyadic decomposition of the discrete wavelet transform (DWT) by applying a high-pass filter $H$ to the original signal $f$ and subsampling by a factor of two to obtain the {\it detail} coefficients and applying a low-pass filter $G$ and subsampling by a factor of two to obtain the {\it approximation} coefficients. Iterating on the approximation coefficients yields a multilevel transform where the $j^{th}$ level extracts information at resolutions between $2^j$ and $2^{j-1}$ pixels. The detail coefficients can be decomposed into horizontal, vertical, and diagonal components when dealing with images.

\autoref{fig:wavelet-transforms} presents an example of a two-level dyadic wavelet transform. The leftmost image is the input image. On the center-left image, the northwest, southwest, and southeast panels represent the {\it detail} coefficients (horizontal, diagonal, and vertical, respectively) at the smallest scale obtained from the high pass filtering of the input image. The northeast panel represents the {\it approximation} coefficients obtained from the low-pass filtering of the input image. On the center-right image, in the northeast panel, the localization of the detail and approximation coefficients is analogous to the decomposition from stage 1; instead, the reference is the approximation from stage 1, not the input image. On the rightmost image, the third decomposition level is obtained from the high pass filtering of the approximation coefficients from stage 2. 

The detail coefficients at stage 1 show details at the 1-2 pixel scale. The detail coefficients at stage 2 are the 2-4 pixel scale details. At stage 3, we can see the detailed coefficients at the 4-8 pixel scale. The approximation coefficients contain everything larger than 8 pixels on the image.

\paragraph{Sobol sensitivity analysis} Let $(X_1,\dots,X_K)$ be independent random variables and $\mathcal{K}= \{1,\dots, K\}$ denote the set of indices. Let $f$ be a model, $X$ an input, and $f(X)$ the model's decision (e.g., the output probability). We denote $f_\kappa = f_\kappa(X_\kappa)$ the partial contributions of the variables $(X_k)_{k\in\kappa}$ to the score $f(X)$. The Sobol-Hoeffding decomposition \cite{hoeffding_class_1992} decomposes the decision score $f(X)$ into summands of increasing dimension
\begin{equation}\label{eq:sobol-hoeffding}
    f(X) = f_\emptyset + \sum_{\kappa\in\mathcal{P}\left(\mathcal{K}\right)\backslash\{\emptyset\}} f_\kappa(X_\kappa),
\end{equation}
Where $f_\emptyset$ denotes the prediction with no features. Under the orthogonality condition  $\forall(u,v)\in\mathcal{K}^2$ such that $u\neq v$, $\mathbb{E}\left[f_u(X_u)f_v(X_v)\right] = 0$, we derive from \autoref{eq:sobol-hoeffding} the variance of the model's score
\begin{equation}\label{eq:variance}
    Var(f(X)) = \sum_{\kappa\in\mathcal{P}\left(\mathcal{K}\right)}Var(f_\kappa(X_\kappa)),
\end{equation}
\autoref{eq:variance} enables us to describe the influence of a subset $\kappa$ of features as the ratio between its own and total variance. This corresponds to the first order {\bf Sobol index} given by 
\begin{equation}\label{eq:sobol-def}
S_\kappa = \frac{Var(f_\kappa(X_\kappa))}{Var(f(X))}.    
\end{equation}
$S_\kappa$ measures the proportion of the output variance $Var(f(X))$ explained by the subset of variables $X_\kappa$ \cite{sobol_sensitivity_1990}. In particular, $S_k$ only captures the {\it direct} contribution of the feature $X_k$ to the model's decision. To capture the indirect effect, due to the effect of $X_k$ on the other variables, {\bf total Sobol indices} $S_{T_k}$ \cite{homma_importance_1996} can be computed as 
\begin{equation}
S_{T_k} = \sum_{\kappa\in\mathcal{P}\left(\mathcal{K}\right),\,k\in\kappa}S_\kappa.
\end{equation}
Total Sobol indices (TSIs) measure the contribution of the $k^{th}$ feature, taking into account both its {\it direct} effect and its {\it indirect} effect through its interactions with the other features. 

\paragraph{Efficient estimation of Sobol indices} As seen from \autoref{eq:sobol-def}, estimating the impact of a feature $k$ on the model's decision requires recording the partial contribution $f_k(X_k)$. This partial contribution corresponds to a {\it forward}. Estimating Sobol indices requires computing variances by drawing at least $N$ samples and computing $N$ forwards to estimate a first-order Sobol index $S_k$ of a single feature $k$. As we are interested in the TSI of a feature $k$, we need to estimate the Sobol index of all sets of features $\kappa\in\mathcal{K}$ such that $k\in\kappa$. To minimize the computational cost of this computation, \cite{fel_look_2021} introduced an efficient sampling strategy based on Quasi-Monte Carlo methods \cite{morokoff_quasi-monte_1995} to generate the $N$ perturbations of dimension $K$ applied to the input and used Jansen's estimator \cite{jansen_analysis_1999} to estimate the TSIs given the models' outputs and the quasi-random perturbations. Their approach requires $N(K+2)$ forwards \cite{fel_look_2021}. 

\begin{figure}[h]
    \centering
    \includegraphics[width = \textwidth]{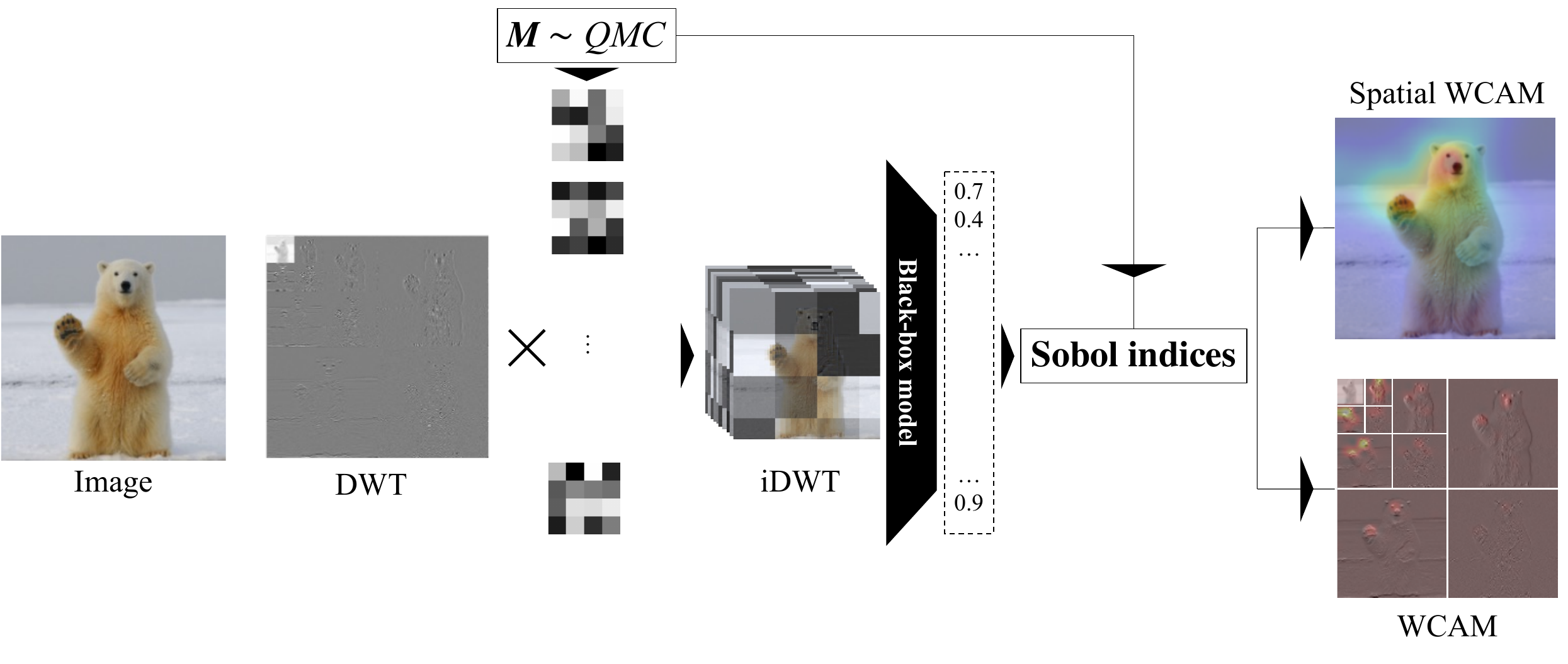}
    \caption{Flowchart of the wavelet scale attribution method (WCAM). \cite{kasmi_assessment_2023} generate a set of perturbed images by perturbing the image's discrete wavelet transform (DWT). Expanding \cite{fel_look_2021}, they apply masks generated from a Quasi Monte-Carlo sequence to the DWT of the image. They evaluate the model on the altered samples reconstructed from the perturbed DWT. They compute the Sobol index for each mask component using the predicted probabilities and the perturbation masks using the Jansen estimator \cite{jansen_analysis_1999}}
    \label{fig:wcam-flowchart}
\end{figure}

To estimate the TSIs, they draw two matrices from a Quasi-Monte Carlo sequence of size $N\times K$ and convert them into perturbations, which they apply to $X$. The perturbated input yields two matrices, $A$ and $B$. $a_{jk}$ (resp. $b_{kj}$) is the element of $A$ (resp. $B$) corresponding to the $k^{th}$ feature and the $j^{th}$ sample. For the $k^{th}$ feature, they define $C^{(k)}$ in the same way as $A$, except that the column corresponding to feature $k$ is replaced by the column of $B$. They then derive an empirical estimator for the Sobol index and TSI as
\begin{equation}
    \hat{S}_k = \frac{\hat{V} - \frac{1}{2N}\sum_{j=1}^N \left[f(B_j) - f\left(C_j^{(k)}\right)\right]^2}{\hat{V}}, \;\;\;\;\;\;\; \hat{S}_{T_k} = \frac{\frac{1}{2N}\sum_{j=1}^N \left[f(A_j) - f\left(C_j^{(k)}\right)\right]^2}{\hat{V}},
\end{equation}
where $f_\emptyset = \displaystyle{
\frac{1}{N}\sum_{j=1}^N f(A_j)
}$ and $\hat{V} = \displaystyle{
\frac{1}{N-1}\sum_{j=0}^N \left[f\left(A_j\right) - f_\emptyset\right]^2
}$. Further implementational details can be found in \cite{fel_look_2021}. \autoref{fig:wcam-flowchart} summarizes the Wavelet sCale Attribution Method of \cite{kasmi_assessment_2023}.

\newpage
\subsection{Decomposing the scales of the PV panels}\label{sec:decomposition}

\begin{figure}[h]
    \centering
    \includegraphics[width=\textwidth]{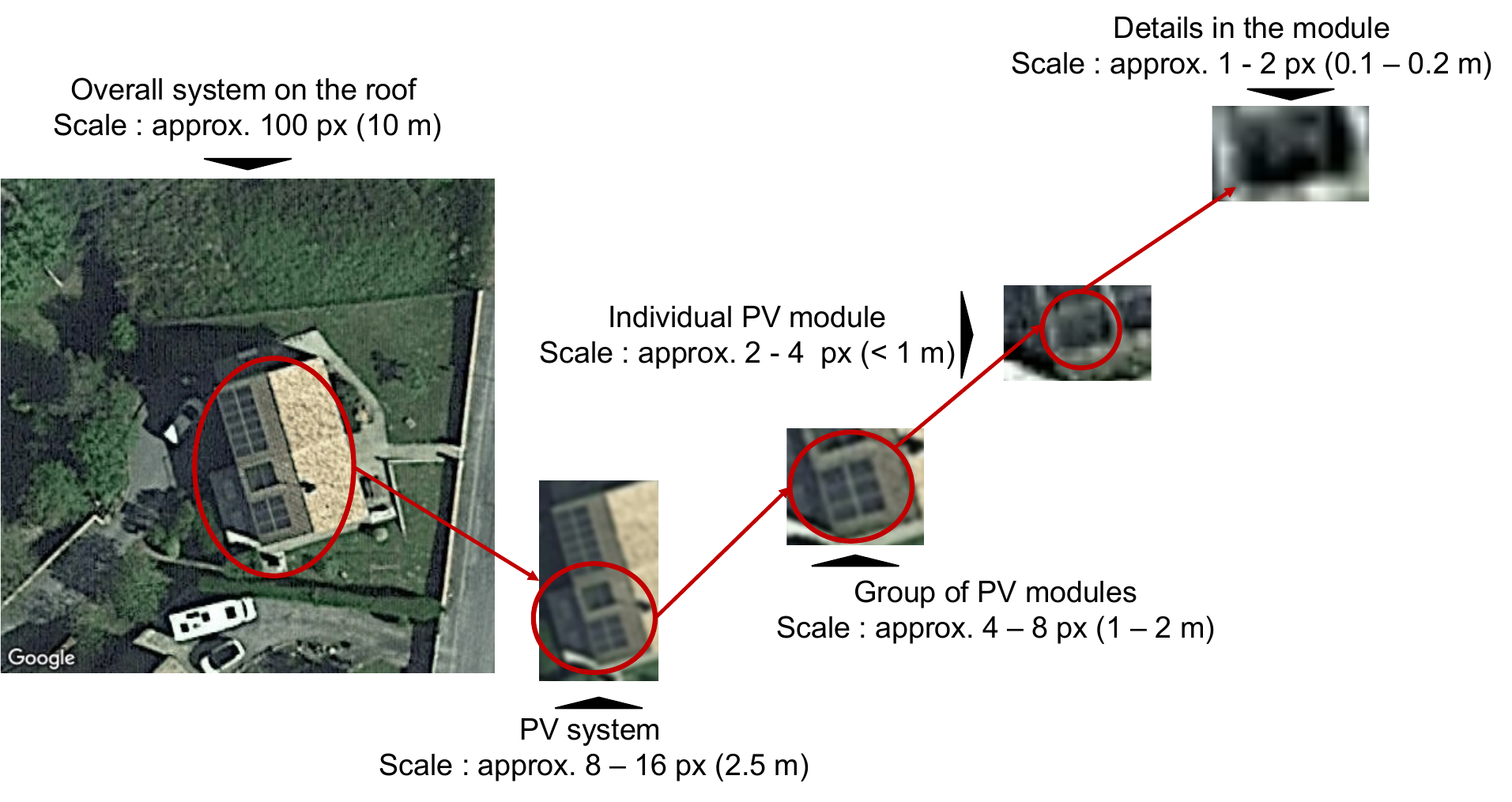}
    \caption{Decomposition of the scales of a PV panel}
    \label{fig:scales_decomposition}
\end{figure}

On orthoimagery, PV panels are located in space and in scale. \autoref{fig:scales_decomposition} shows an example. Depending on the scale of interest, the PV panel will show different characteristics. At the smallest scales (i.e., high frequencies), the PV panel corresponds to small details within the individual PV modules. On the opposite, if we consider the PV system as a whole, it has a size of about 10m, i.e., 100 pixels in this image. 

The WCAM enables us to see the important scales for the prediction and their location. It lets us know what the model sees from the panel: the overall system, the cluster, or individual modules.

Besides, it also illustrates the impact of background noise on the detection. If small scales not located on the panel are important for the prediction, it indicates that the model relies on background noise for the prediction. For example, the image of \autoref{fig:scales_decomposition} could be the grass around the house.

\subsection{Some examples}

\begin{figure}[h]
    \centering
    \includegraphics[width = \textwidth]{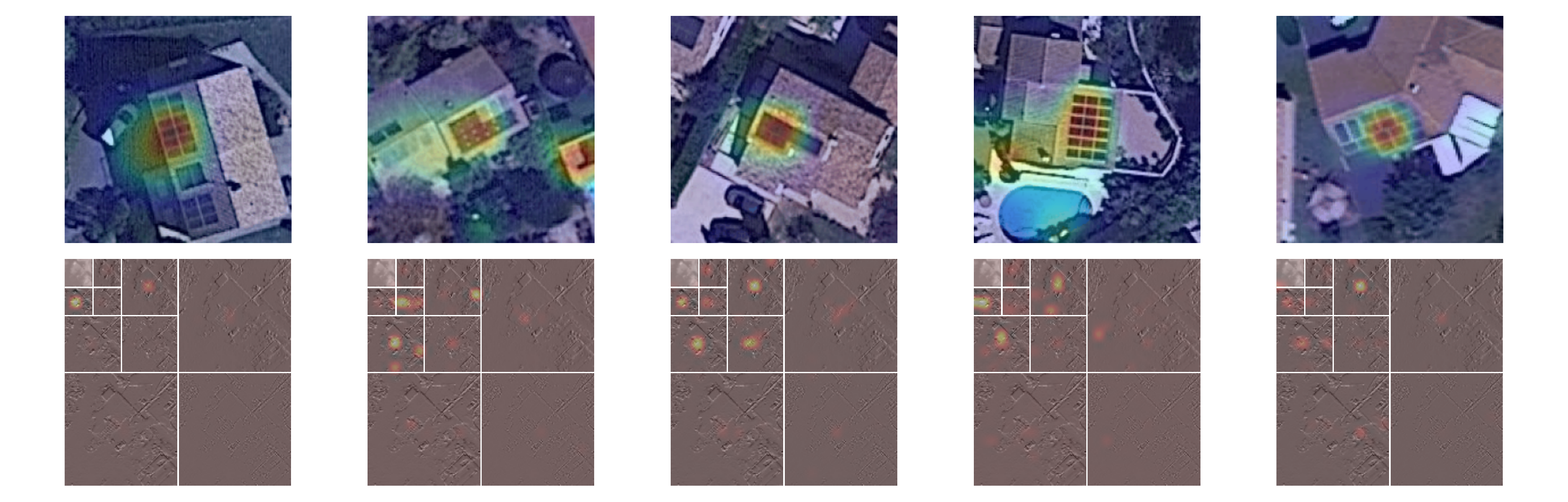}
    \caption{Decomposition in the space-scale domain of PV panel predictions}
    \label{fig:true_pos_example}
\end{figure}

\autoref{fig:true_pos_example} presents some examples of predictions made by a PV classifier. We can see that for our illustration image (leftmost image on \autoref{fig:true_pos_example}), the grass, but also the shape of the garden on the upper left of the image, play a small role in the prediction of the PV panel. On the other hand, in the two subsequent images, the prediction relies on factors at different scales located around the PV panel. 

The WCAM lets us see that when a model focuses on a PV panel, its focus can be decomposed into different scales. Besides, these scales do not always have the same importance. For instance, the importance of the 1-2 pixel details is more important on the third image from the left than for the other images. It could be because the panel is frameless, so the model has to rely on other factors, as this one is missing for this image.

\section{Data augmentation strategies}

\subsection{Description of the data augmentations}\label{sec:augmentation-plot}

\paragraph{AugMix \cite{hendrycks_augmix_2020}} The data augmentation strategy "Augment-and-Mix" (AugMix) consists of producing a high diversity of augmented images from an input sample. A set of operations (perturbations) to be applied to the images are sampled, along with sampling weights. The image resulting $x_{aug}$ is obtained through the composition $x_{aug} = \omega_1 op_1 \circ \dots \omega_n op_n (x) $ where $x$ is the original image. Then, the augmented image is interpolated with the original image with a weight $m$ that is also randomly sampled. We have $x_{augmix} = mx + (1-m)x_{aug}$. 

\paragraph{AutoAugment \cite{cubuk_autoaugment_2019}} This strategy aims at finding the best data augmentation for a given dataset. The authors determined the best augmentations strategy $S$ as the outcome of a reinforcement learning problem: a controller predicts an augmentation policy from a search space. Then, the authors train a model, and the controller updates its sampling strategy $S$ based on the train loss. The goal is that the controller generates better policies over time. The authors derive optimal augmentation strategies for various datasets, including ImageNet \cite{russakovsky_imagenet_2015}, and show that the optimal policy for ImageNet generalizes well to other datasets.

\paragraph{RandAugment \cite{cubuk_randaugment_2019}} This strategy's primary goal is to remove the need for a computationally expansive policy search before model training. Instead of searching for transformations, random probabilities are assigned to the transformations. Then, each resulting policy (a weighted sequence of $K$ transformations) is graded depending on its strength. The number of transformations and the strength are passed as input when calling the transformation. 

\paragraph{Blurring} We apply a nonrandom Gaussian blur to the image. The value is set by comparing visually Google and IGN images and trying to remove details from Google images that are not visible on IGN images. After a manual inspection, we set the blur level so that the details at 10-20cm scale are discarded from the image. It corresponds to a blurring value $\sigma = 2.$ in the {\tt ImageFilter.GaussianBlur} method of the PIL library. 

\paragraph{Blurring + Wavelet perturbation (WP)} We first blur the image. Then, for each color channel, we compute the dyadic wavelet transform of the image and randomly perturb the coefficients (we randomly set some coefficients to 0). The set of coefficients set to 0 is determined with uniform sampling. This results in a random perturbation that removes information for some precise scales and locations. We then reconstruct the image from its perturbed wavelet coefficients. For each call, 20\% of the coefficients are canceled. This value balances between the loss of information and the input perturbation. We perturb each color channel independently.


\subsection{Plots}\label{sec:plot-augmentation}

\begin{figure}[h]
    \centering
    \includegraphics[width = \textwidth]{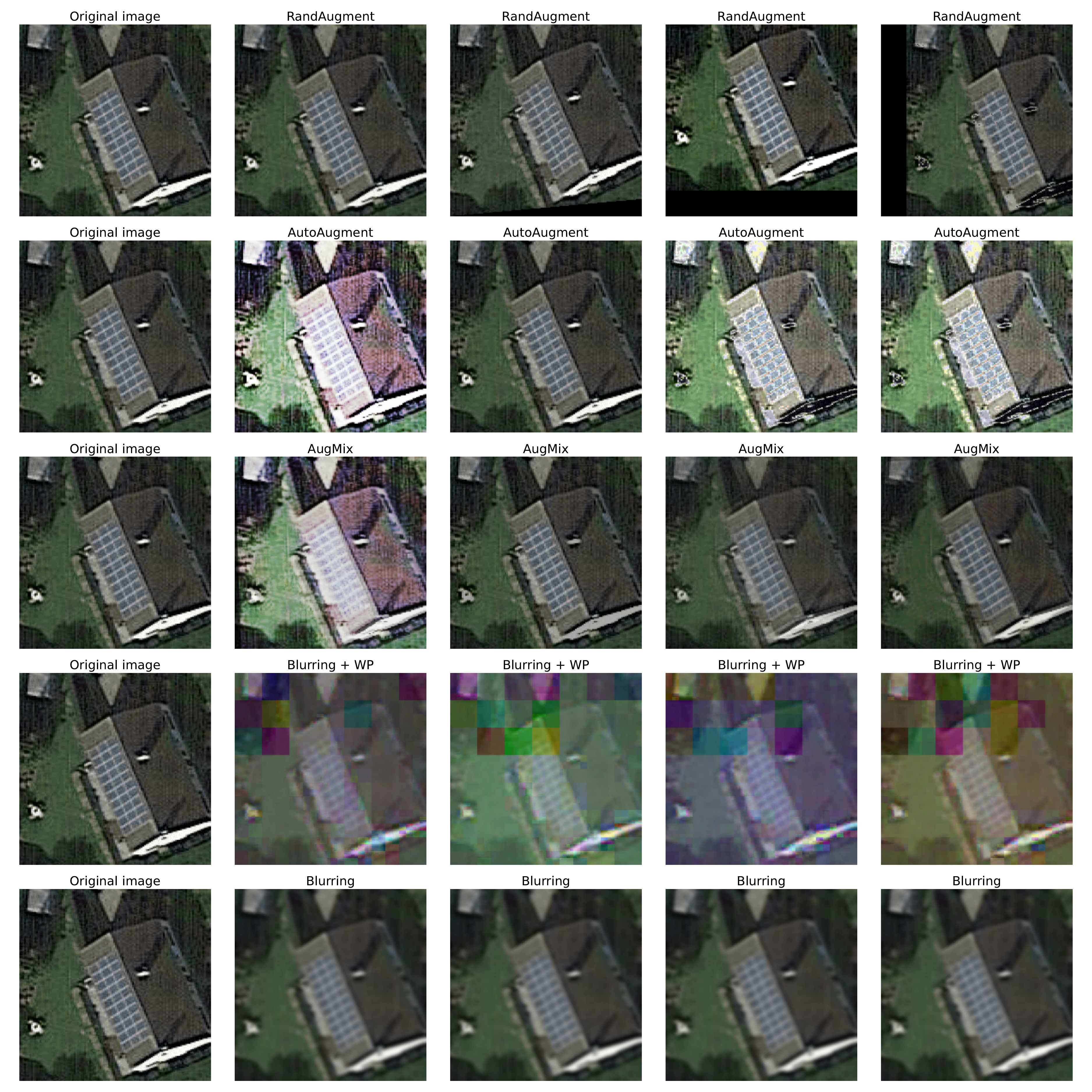}
    \caption{Visualization of the different data augmentation techniques implemented in this work.}
    \label{fig:augmentations}
\end{figure}

\autoref{fig:augmentations} plots examples of the different data augmentations implemented in this work. Along with these augmentations, we apply random rotations, symmetries, and normalization to the input during training. At test time, we only normalize the input images.

\newpage
\section{Complementary results}\label{sec:complementary-results}

\subsection{Training results on the Google test set}

\paragraph{Training results} \autoref{tab:mitigation-source} reports the training results of our methods on the source (Google) test set. We can see that our spectral method performs slightly less than the other methods on the source dataset.

  \begin{table}[h]
\small
  \centering
  \vspace{0mm}\caption{\textbf{F1 Score} and decomposition in true positives, true negatives, false positives, and false negatives for models trained on Google images with different strategies to mitigate the sensitivity to acquisition conditions. Evaluation computed on the Google (source) dataset.}\label{tab:mitigation-source}
  \resizebox{\textwidth}{!}{\begin{tabular}{r c c c c c}
  \toprule
   
   & F1 Score ($\uparrow$) & True positives & True negatives & False positives & False negatives \\
   \midrule
    ERM \cite{vapnik_nature_1999} & 0.98 & 1891 & 2355 & 36 & 39 \\
    AutoAugment \cite{cubuk_autoaugment_2019} &  0.98 & 1906 & 2340 & 51 & 24 \\
    AugMix \cite{hendrycks_augmix_2020} & 0.98 & 1894 & 2354 & 37 & 36 \\
    RandAugment \cite{cubuk_randaugment_2019} & 0.98 & 1907 & 2342 & 49 & 23 \\
    Blurring  & 0.82 & 1636 & 1958 & 433 & 294\\
    Blurring + WP &  0.90 & 1798 & 2135 & 256 & 132\\
  \bottomrule
  \end{tabular}}
  \end{table}

\subsection{Probability shift} 

\begin{figure}[h]
    \centering
    \includegraphics[width = .7\textwidth]{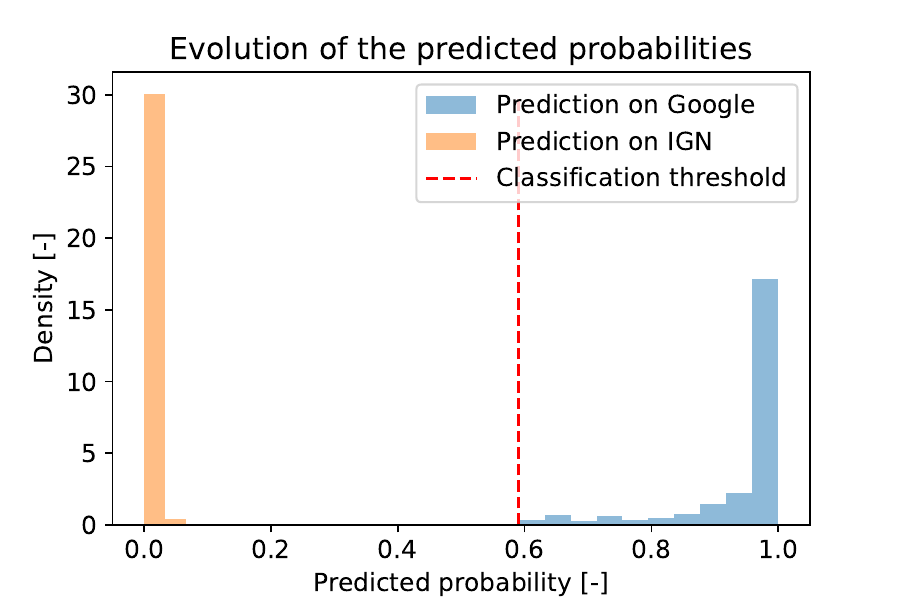}
    \caption{Evolution of the predicted probabilities for images depicting a PV panel on the Google test set and the corresponding images on the IGN test set. The predicted probability completely flips over when the model no longer recognizes the PV panel.}
    \label{fig:prob-shift}
\end{figure}

\autoref{fig:prob-shift} shows the change in predicted probabilities for positive Google images. We can see that when the classifier no longer recognizes the PV panel, the probability shift is large, suggesting that the important factor for prediction disappeared from the image. These results provide more systematic evidence of the qualitative pattern highlighted by \autoref{fig:wcam-example}: if a critical component is no longer depicted due to the change in the acquisition condition, then the model no longer sees a PV panel (despite having other information about it).

\subsection{Understanding false positives} 

\begin{figure}[t]
    \centering
    \includegraphics[width = \textwidth]{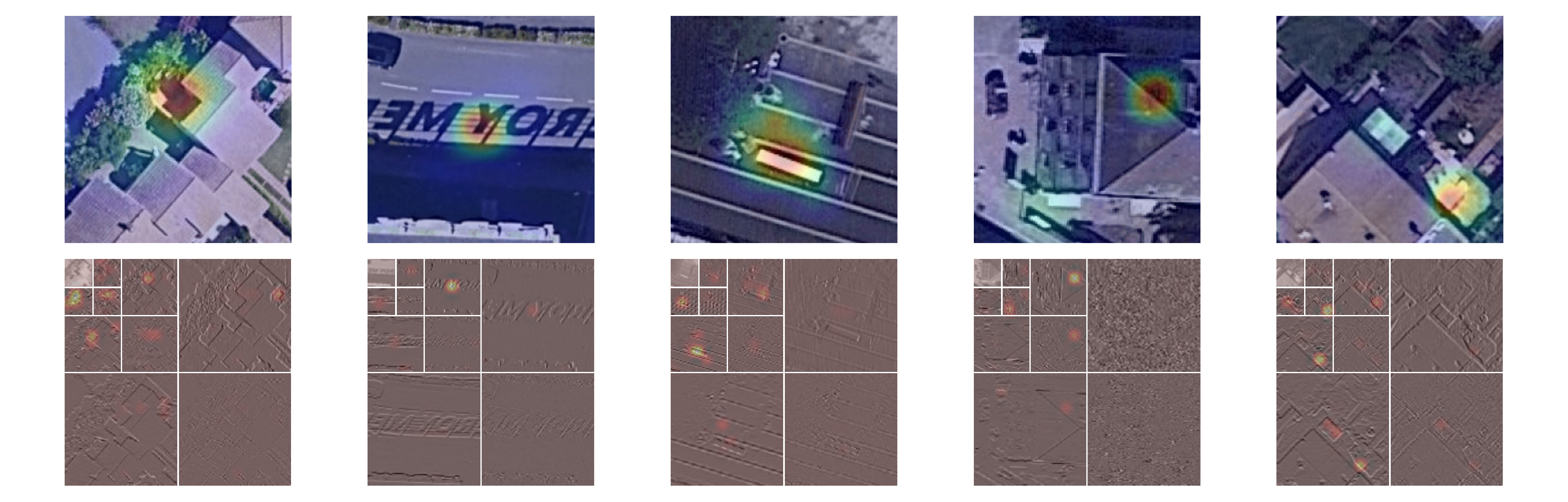}
    \caption{Examples of false positives on IGN and corresponding WCAM.}
    \label{fig:false-positives}
\end{figure}

On \autoref{fig:false-positives}, we plot some examples of false positives and their associated WCAM. We can see that the WCAM helps us understand why the model misleads an element of the image with a PV panel.

In the leftmost image, we can see that the shape of the roof mainly causes the false alarm, which is reminiscent of the delineation of a PV system on a roof. On the other hand, in the fourth image (from the left), multiple factors at different scales lead to a wrong prediction.

\end{document}